\documentclass[a4paper, 11 pt, conference]{ieeeconf}  
\usepackage{amsmath}
\IEEEoverridecommandlockouts                              
\makeatletter
\let\NAT@parse\undefined
\makeatother
\usepackage[dvipsnames]{xcolor}

\newcommand*\linkcolours{ForestGreen}

\usepackage{times}
\usepackage{graphicx}
\usepackage{amssymb}
\usepackage{textcomp,gensymb}
\usepackage{amsmath}
\usepackage{breakurl}

\usepackage{url,hyperref}
\hypersetup{
colorlinks,
linkcolor=\linkcolours,
citecolor=\linkcolours,
filecolor=\linkcolours,
urlcolor=\linkcolours}

\usepackage{algorithm}
\usepackage{algorithmic}
\usepackage{resizegather}

\usepackage[labelfont={bf},font=small]{caption}
\usepackage[none]{hyphenat}

\usepackage{mathtools, cuted}

\usepackage[noadjust, nobreak]{cite}

\usepackage{tabularx}
\usepackage{amsmath}

\usepackage{float}

\usepackage{pifont}

\newcolumntype{Y}{>{\centering\arraybackslash}X}

\usepackage[]{placeins}

\usepackage[square,sort,comma,numbers]{natbib}
\usepackage{placeins}

\usepackage{tikz}

\usepackage[framemethod=tikz]{mdframed}

\usepackage{afterpage}

\usepackage{stfloats}

\usepackage{atbegshi}
\newcommand{\handlethispage}{}
\newcommand{\discardpagesfromhere}{\let\handlethispage\AtBeginShipoutDiscard}
\newcommand{\keeppagesfromhere}{\let\handlethispage\relax}
\AtBeginShipout{\handlethispage}

\usepackage{comment}
\usepackage{mathtools, esdiff}

\title{\LARGE \bf
Transformer++
}

\author{ \parbox{3 in}{\centering Prakhar Thapak\\
         Machine Learning \& AI Team\\
         American Express\\
         {\tt\small prakhar.thapak@aexp.com}}
         \hspace*{ 0.5 in}
         \parbox{3 in}{ \centering Prodip Hore\\
        Machine Learning \& AI Team\\
         American Express\\
         {\tt\small prodip.hore@aexp.com}}
}

\usepackage{graphicx}

\begin{document}

\maketitle
\thispagestyle{empty}
\pagestyle{empty}

\begin{abstract}
Recent advancements in attention mechanisms have replaced recurrent neural networks and its variants for machine translation tasks. Transformer using attention mechanism solely achieved state-of-the-art results in sequence modeling. Neural machine translation based on the attention mechanism is parallelizable and addresses the problem of handling long-range dependencies among words in sentences more effectively than recurrent neural networks. One of the key concepts in attention is to learn three matrices, query, key, and value, where global dependencies among words are learned through linearly projecting word embeddings through these matrices. Multiple query, key, value matrices can be learned simultaneously focusing on a different subspace of the embedded dimension, which is called multi-head in Transformer. We argue that certain dependencies among words could be learned better through an intermediate context than directly modeling word-word dependencies. This could happen due to the nature of certain dependencies or lack of patterns that lend them difficult to be modeled globally using multi-head self-attention. In this work, we propose a new way of learning dependencies through a context in multi-head  using convolution.  This new form of multi-head attention along with the traditional form achieves better results than Transformer on the WMT  2014  English-to-German and English-to-French translation tasks. We also introduce a framework to learn POS tagging and NER information during the training of encoder which further improves results achieving a new state-of-the-art of 32.1 BLEU, better than existing best by 1.4 BLEU, on the  WMT 2014 English-to-German and 44.6 BLEU, better than existing best by 1.1 BLEU, on the WMT 2014 English-to-French translation tasks. We call this Transformer++.

\end{abstract}

\section{\textbf{INTRODUCTION}}

Neural machine translation addresses the problem of translating one language into another language using neural networks. Typical encoder-decoder models read the entire sentence from the source language and encode it to a context vector. The task of the Decoder is to read this encoded vector and emit the words in the target language in a sequential fashion. Recurrent neural network and their variants have typically achieved state-of-the-art results in machine translation tasks \cite{DBLP:journals/corr/BahdanauCB14,DBLP:journals/corr/ChoMGBSB14,DBLP:journals/corr/ChungGCB14,hochreiter1997long,DBLP:journals/corr/ShazeerMMDLHD17}. Attention mechanism along with recurrent neural network has found usefulness to learn long term dependencies as it allows to establish alignment between words in input and output sequences directly irrespective of the distances \cite{DBLP:journals/corr/BahdanauCB14,DBLP:journals/corr/KimDHR17,parikh2016decomposable,DBLP:journals/corr/LuongPM15,DBLP:journals/corr/ChorowskiBSCB15,DBLP:journals/corr/SukhbaatarSWF15}. Recurrent neural networks are sequential in nature, which makes them difficult to parallelize and exploit GPU. It is also difficult to relate words separated by long distances using a recurrent neural network \cite{DBLP:journals/corr/GehringAGYD17,hochreiter2001gradient,NIPS2016_6295,DBLP:journals/corr/KalchbrennerESO16}.

\vspace{1.5mm}

Transformer\cite{nips} introduced in $2017$ achieved state-of-the-art results on machine translation task using only attention mechanism eschewing recurrent neural networks. This was significant given attention mechanism based algorithms are parallelizable and can exploit GPU. Attention mechanism has proven to be useful as one can model dependencies of words in sequences irrespective of the distance between them.

\vspace{1.5mm}

In attention mechanism, three matrices query, key and value, are learned through backpropagation where global dependencies among words are learned through linearly projecting word embeddings through these matrices. Multiple query, key and value matrices can be learned through different representations, which allows the model to attend to information from different representation subspaces. This is called as multi-head attention in Transformer \cite{nips}.  

\vspace{1.5mm}
   
Along with learning global dependencies among words in self-attention \cite{nips}, we argue that a machine translation task may also benefit from learning dependencies through an intermediate context than directly modeling word-word dependencies. This could happen due to the nature of certain dependencies that lend them difficult to  be modeled globally in multi-head of self-attention or due to lack of patterns that makes these dependencies not obvious when modeled directly among words. In this work, we propose a new way of learning these dependencies in multi-head of self-attention \cite{nips} using convolution\cite{DBLP:journals/corr/KalchbrennerESO16,DBLP:journals/corr/KaiserGC17,DBLP:journals/corr/Chollet16a}. We explicitly model word-context dependencies, that is, which words are more relevant to a context. The context could be the summary of an entire sentence. This can be thought of as a novel way of computing alignments scores in the attention mechanism. In Transformer, the attention scores are being calculated modeling word-word dependencies, here we explicitly model word-context dependencies in multi-head. For example, consider the sentence `There is a fire in the mountain'. For the query `fire', we want a high attention score for the word `mountain.' It is observed using proposed multi-head attention they are closer in the embedded space, and we believe this helps learn these kinds of dependencies better. Co-occurrences of words, such as  `fire' and `mountain' may not be usual in a corpus; hence globally modeling as dependencies between words may be difficult. We also demonstrate that using POS tagging and NER information during the training of encoder helps in translation tasks and further improves results. Using POS and NER information has shown better results in recurrent neural networks in the past \cite{DBLP:journals/corr/abs-1708-00993,li-etal-2018-named}. We incorporate that in attention based encoder. To the best of our knowledge, this achieves a new state-of-the-art in machine translation tasks on the publicly available WMT 2014 English-to-German and WMT 2014 English-to-French translation tasks. We call it Transformer++. 

\vspace{1.5mm}

\section{\textbf{Background}}

Transformer achieved state-of-the-art results in neural machine translation without using recurrent neural networks and its variants. Sequential nature and difficulty in modeling long-range dependencies were drawbacks of recurrent neural networks and related algorithms \cite{hochreiter2001gradient}.  There has been work to reduce sequential computational requirements in machine translation tasks, and most of them are based on using convolutional neural networks as building blocks \cite{DBLP:journals/corr/GehringAGYD17,NIPS2016_6295,DBLP:journals/corr/KalchbrennerESO16}. Still modeling long-range dependencies is a challenge as the computational requirement grows linearly or logarithmically in these algorithms \cite{hochreiter2001gradient,nips}. In \cite{DBLP:journals/corr/abs-1901-10430} using a simple lightweight convolution neural network, having a small number of parameters,  they achieved comparable results to the Transformer, and using a dynamic convolutional network achieved better results than the Transformer. These models are also reported to be computationally faster than the Transformer. This shows that convolutional neural networks based algorithms also hold promise in neural translation tasks and also competitive in computation cost.  In this work, we focus on improving the Transformer architecture which is based on self-attention and introduce the concept of modeling words-context dependencies in multi-head along with learning POS tagging and NER information. This achieves a new state-of-the-art of $32.1$ BLEU on the WMT 2014 English-to-German and $44.6$ BLEU on the WMT 2014 English-to-French translation. 

\section{\textbf{Model Architecture}}

\begin{figure}[tbp]
\centering
\includegraphics[width=9cm,height=7cm]{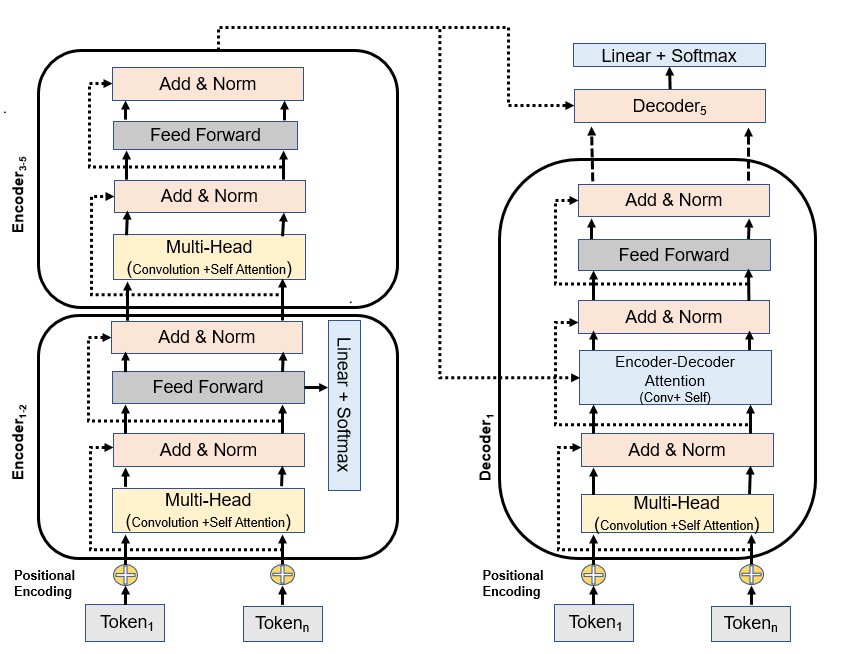}
\caption{Transformer++ model architecture}
\label{whole model}
\end{figure}

We use an encoder-decoder architecture as shown in figure[\ref{whole model}] for a sequence to sequence modeling following the architecture presented in Transformer \cite{nips}. We propose a new way to calculate attention function to learn both proposed new multi-head and traditional multi-head simultaneously. For a given $H$ (total number of multi heads), we use self-attention in $H/2$ heads to capture global dependencies and use Convolution based attention in $H/2$ heads to capture dependencies through context. This explicitly models word-context dependencies using Convolution in $H/2$ heads. We demonstrate that word-context dependencies will complement traditional multi-head. Traditional multi-head can be thought of explicitly modeling word-word dependencies,  linearly projecting words embeddings through three matrices query, key, and value.

\vspace{1.5mm}

\subsection{\textbf{Encoder Stack}}
\begin{figure}[tbp]
\centering
\includegraphics[width=1\columnwidth]{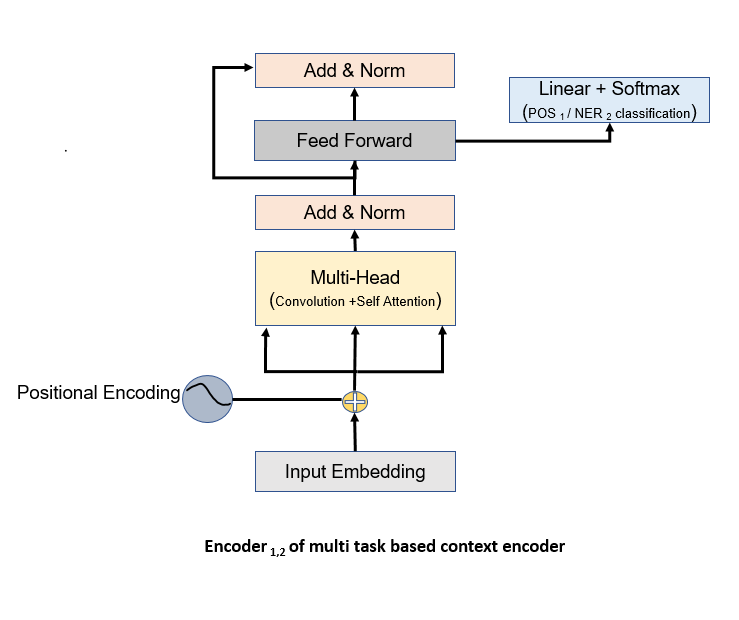}
\caption{Base Encoder}
\label{encoder structure}
\end{figure}

The encoder is composed of a base encoder followed by the standard encoder proposed as in Transformer \cite{nips}. The Base encoder is composed of a stack of $N = 2$ layers to accommodate the learning of POS tagging and NER, which we will explain later. These two encoders have three sub-layers. The first is a multi-head attention mechanism (self-attention + proposed context-word attention) layer, while the second layer is a simple position-wise fully connected feed-forward network followed by a linear + softmax layer to train on POS/NER labels as shown in figure[\ref{encoder structure}]. We use residual connection \cite{he2016deep} around each of the two multi-head and feed-forward layers followed by layer normalization \cite{ba2016layer}. The idea is to use multi-task learning to train encoder\textsubscript{1}  for part of speech tagging as well as for machine translation. This allows the encoder layer to attending the syntactic functions of the input language. The encoder\textsubscript{2} is again trained in a multi-task manner. This encoder, along with machine translation, is used for learning named entity recognition forcing the encoder to attend to information extraction. The output from encoder\textsubscript{2} is connected to a minimum of $1$ additional encoder, which is the same as proposed in the Transformer. This encoder takes the output produced by the base encoders to produce context encoder output. The labels for part of speech tagging and named entity recognition were obtained using the Spacy library. 

\vspace{1.5mm}

\subsection{\textbf{Decoder Stack}}
We start with the decoder architecture in Transformer\cite{nips} and propose changes in the multi-head attention, which is a combination of self-attention and convolution as shown in figure[\ref{whole model}]. The decoder also consists of $N$ identical layers but doesn't have the extra linear and softmax layer used for POS and NER classification in the encoder. A multi-Head attention is also employed between the output of the final encoder layer and the output of each decoder layer after self-attention and feed-forward layer as in \cite{nips}. We use residual connections and layer normalization as employed in the encoder. The self-attention sub-layer in the decoder uses masking to avoid looking at the subsequent tokens. This masking ensures that the predictions for position $i$ can depend only on the tokens less than $i$.

\vspace{1.5mm}

\subsection{\textbf{Attention}}

We use the scaled dot-product attention that takes care of hard softmax problem \cite{nips} along with the proposed context–word attention.

\vspace{1.5mm}
\subsubsection{\textbf{Scaled Dot-Product Attention}}
\vspace{0.1mm}
We used the Scaled Dot-Product Attention in $H/2$ heads.
The input consists of queries (Q) and keys (K) of dimension $d\textsubscript{k}$, and values (V) of dimension $d\textsubscript{v}$\cite{nips}.

\begin{multline}
Attention(K,Q,V)=softmax(QK^T/\sqrt{d_{k}})V
\end{multline}

\vspace{1.5mm}

\subsubsection{\textbf{Convolution Attention}}
\begin{figure}[tbp]
\centering
\includegraphics[width=1\columnwidth]{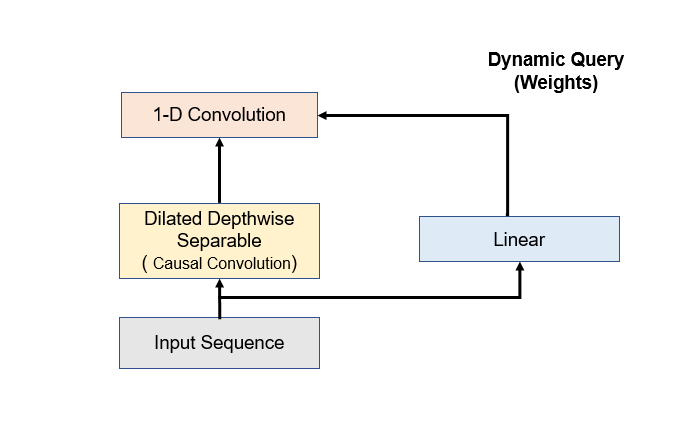}
\caption{Convolution Attention}
\label{convolution structure}
\end{figure}

We propose two modules for convolution attention. First is the adaptive sequence module, which captures the local context for each word. The other being an adaptive query module, which captures the entire context for an input sequence as shown in figure[\ref{convolution structure}]. Convolution has a fixed context window that helps in determining the importance of a word in a local context. The input sequence is first passed through a dilated causal convolution kernel\cite{DBLP:journals/corr/KalchbrennerESO16,DBLP:journals/corr/KaiserGC17,DBLP:journals/corr/Chollet16a} which is depth-wise separable. Causal convolution helps in ensuring that there is no information leak while the weights across temporal dimension $F$ are softmax normalized so that a word representation is a weighted average of the words in a context. We use dilation \cite{DBLP:journals/corr/YuK15} to increase the receptive field of the context window. Convolution being depth-wise helps in reducing the parameters from $d\textsuperscript{2}F$ to $dF$ where $d$ is the dimension in word representation and $F$ is the filter size. The output of the dilated depth-wise casual convolution is a token representation having a local context.

\vspace{1.5mm}
Adaptive sequence module calculates the following local context for $t\textsuperscript{th}$ element in a sequence $S$ across channels $d$.

\begin{multline}
LocalConv(S,W\textsubscript{A},t,d)= \\
Conv(S,softmax(W\textsubscript{A}),t,d) \;\; ;W\textsubscript{A} \in \mathbb{R} ^ {F \times d}
\end{multline}

\vspace{1.5mm}
The output from the adaptive sequence module represents the local context of a word across context $F$.

\vspace{2.5mm}
Adaptive query module: The input sequence $S$ is passed through linear layers, $W\textsuperscript{S}$ and $W\textsuperscript{Q}$, to obtain the Query matrix, which gets convolved with the output of the adaptive sequence module. This is conceptually similar to obtaining context using recurrent neural networks \cite{DBLP:journals/corr/BahdanauCB14}, hence the output of the linear layer after multiplying with $W\textsuperscript{Q}$ is softmax normalized to obtain the weighted sum of the projection from the linear layer using $W\textsuperscript{S}$. For a sequence $S$ with length $T$ the dynamic query is given by :

\begin{multline}
Query= \sum_{t=1}^T {(SW^S)_t \; softmax_t(S \times W^Q)} \,;\\ W^Q\in \mathbb{R} ^d \; \& \; W^S\in \mathbb{R}^{d \times d }
\end{multline}

\begin{multline}
\small DynamicConv(S,t,d)=\\
Conv1D(LocalConv,Query)
\end{multline}

In equation [4] above, Conv1 is simply computing the dot product between LocalConv and Query. In new multi-head, LocalConv is the word representation after convolution and Query represents the context of the entire sentence; hence attention score being computed in word-context relations. LocalConv is conceptually similar to the key matrix and DynamicConv as the value matrix in self-attention.

\subsection{\textbf{Multi-Head Attention}}

For a given $H$ (total number of multi-head), we use  $H/2$ self-attention heads to capture global dependencies directly among words and use $H/2$ Convolution based heads to capture dependencies among words through a context. We demonstrate this novel way of capturing word-context dependencies will complement traditional multi-head and help model certain dependencies among words more meaningfully though a context rather directly. For $W^O\in \mathbb{R} ^{d \times d}$. 

\begin{multline}
MultiHead(S)=Concat(head\textsubscript{1}...head\textsubscript{H})W^O 
\end{multline}

\begin{gather}
head\textsubscript{i,1:H/2}=SelfAttn(QW\textsubscript{i}^Q,KW\textsubscript{i}^K,VW\textsubscript{i}^V)\nonumber \\
head\textsubscript{H/2:H}=DynamicConv(S,t,d)\nonumber
\end{gather}

\vspace{1.5mm}
Multi-head attention allows the model to attend  different subspace of the embedded dimension simultaneously. In order to learn both the attention jointly,  the output from the convolution attention and the self-attention, is concatenated and passed through a linear layer as in \cite{nips}. In Table III-D, we compare the complexities for different layer types where $n$ is the sequence length, $d$ is the embedding dimension and $f$  refers to kernel size in convolution. Introducing word-context attention using depthwise separable convolution is of comparable cost.

\vspace{5mm}
\begin{table}
\LARGE
\label{complexity}
\scalebox{0.6}{%

\begin{tabular}{ |p{3.5cm}|p{2.7cm}|p{2.7cm}| p{2.7cm}| }
\hline
\textbf{Layer Type}  & \textbf{Complexity per Layer} & \textbf{Sequential Operations} & \textbf{Minimum Path Length} \\
\hline
Self-Attention & $O(n^2.d)$ & $O(1)$ & $O(1)$\\
\hline
Recurrent & $O(n.d^2)$  & $O(n)$  & $O(n)$ \\
\hline
Convolution & $O(f.n.d^2)$ & $O(1)$  & $O(\log\textsubscript{f}(n))$ \\
\hline
Depthwise Separable Convolution & $O(f.n.d)$ & $O(1)$ &$O(\log\textsubscript{f}(n))$ \\
\hline
\end{tabular}}
\caption*{Table III-D}
\end{table}

\section{\textbf{Datasets and Training}}

We used the WMT 2014 English-German dataset and WMT 2014 English - French dataset for training while validating on newstest2013 and test on newstest2014. We replicate the setup of Transformer for training and used Adam optimizer \cite{DBLP:journals/corr/KingmaB14} with the learning rate presented in Transformer. Tokens were encoded using byte-pair encoding \cite{DBLP:journals/corr/BritzGLL17}. Sentence pairs having same sequence length were batched together for faster computation.

\vspace{2.5mm}
We used a dropout \cite{JMLR:v15:srivastava14a} of $0.25$ on the WMT En-De and $0.15$ on the WMT En-Fr along with residual dropout and token embedding dropout with value $0.10$. We use DropConnect \cite{wan2013regularization} as a regularizer for convolution operation. We train the WMT models on $4$ GPUs for a total of $320K$ steps. We accumulate the gradient for $10$ steps before applying an update \cite{ott2018scaling}. The encoder and decoder both have $N=5$ blocks to match the parameters of the Transformer with encoder\textsubscript{1,2} trained on POS and NER in a multi-task manner. Both encoder and decoder have kernel size $F=3,5,7,11,15$ for each block respectively and $H=16$. We used stochastic weight averaging for the last $10$ checkpoints with beam search of $5$ and length penalty $\alpha=0.5$\cite{wu2016google}.

\section{\textbf{Results and conclusions}}

\vspace{5mm}
\scalebox{0.5}{%
\LARGE
\label{results}
\begin{tabular}{ |p{4cm}|p{3cm}|p{3cm}| p{3cm}| }
\hline
\multicolumn{4}{|c|}{\textbf{Performance of WMT on newstest2014 tests}} \\
\hline
\hline
\textbf{Model}  & \textbf{Param} & \textbf{WMT En-De} & \textbf{WMT En-Fr} \\
\hline
Transformer \cite{nips} & 213M & $28.4$ & $41.8$\\
\hline
Scaling Neural Machine Translation\cite{ott2018scaling} & 210M & $29.3$ & $43.2$\\
\hline
Lightweight and Dynamic convolution\cite{DBLP:journals/corr/abs-1901-10430} & 213M & $29.7$ & $43.2$\\
\hline
Data Diversification \cite{nguyen2019data} & 209M & $\textbf{30.7}$ &$\boldsymbol{-}$ \\
\hline
MUSE\cite{zhao2019muse} & $\boldsymbol{-}$ & $29.9$ & $\textbf{43.5}$ \\
\hline
Transformer++ & 212M & $\textbf{32.1}$ & $\textbf{44.6}$ \\
\hline
\end{tabular}}

\vspace{17mm}
In Table[\ref{results}],  we compare the results of our algorithm with other relevant state-of-the-art algorithms reported on the  WMT 2014 English-German dataset and WMT 2014 English-French dataset. This new form of multi-head attention along with the traditional form achieves better results than Transformer by $1.8$ BLEU score on the  WMT 2014 English-to-German and $1.9$ BLEU on the WMT 2014 English-to-French translation tasks. Training with POS tagging and NER information further improves results achieving new state-of-the-art of $32.1$ BLEU, better than existing best \cite{nguyen2019data} by $1.4$ BLEU, on the  WMT 2014 English-to-German and $44.6$ BLEU, better than the existing best \cite{zhao2019muse} by $1.1$ BLEU, on the WMT 2014 English-to-French translation tasks. To get some insights, we computed the cosine similarity between the embedding of the words `fire' and `mountain' in the example sentence `There is a fire in the mountain', where we anticipated word-context attention might help and bring `fire' and `mountain' closer in the embedded space. In self-attention, the cosine similarity is 0.21 while in the proposed method, computed from the embedding in equation[2], it is 0.39 indicating they came closer.

\vspace{2mm}

\vspace{5mm}

In this work, we proposed novel ways of creating word-context dependencies in multi-head self-attention demonstrating it complements traditional form of multi-head. We believe this will help the research community to explore other methods to learn dependencies in multi-head self-attention for neural machine translation tasks. Incorporating POS tagging and NER information in the training of encoder further improves results. 
To the best of our knowledge, it is the new state-of-the-art, and we call it Transformer++. 

In the future, we plan to study more thoroughly the usefulness of various components in the encoder-decoder architecture, such as the optimal number of encoder and decoder stacks required, orthogonal value created from the proposed new way of creating multi-head attention, sensitivity to parameters,  and explore learning word dependencies more effectively in multi-head attention for neural machine translation tasks.

\medskip
\bibliography{ref} 

\begin{thebibliography}{10}

\bibitem{DBLP:journals/corr/BahdanauCB14}
Dzmitry Bahdanau, Kyunghyun Cho, and Yoshua Bengio.
\newblock Neural machine translation by jointly learning to align and
  translate.
\newblock In Yoshua Bengio and Yann LeCun, editors, {\em 3rd International
  Conference on Learning Representations, {ICLR} 2015, San Diego, CA, USA, May
  7-9, 2015, Conference Track Proceedings}, 2015.

\bibitem{DBLP:journals/corr/ChoMGBSB14}
Kyunghyun Cho, Bart van Merrienboer, {\c{C}}aglar G{\"{u}}l{\c{c}}ehre, Fethi
  Bougares, Holger Schwenk, and Yoshua Bengio.
\newblock Learning phrase representations using {RNN} encoder-decoder for
  statistical machine translation.
\newblock {\em CoRR}, abs/1406.1078, 2014.

\bibitem{DBLP:journals/corr/ChungGCB14}
Junyoung Chung, {\c{C}}aglar G{\"{u}}l{\c{c}}ehre, KyungHyun Cho, and Yoshua
  Bengio.
\newblock Empirical evaluation of gated recurrent neural networks on sequence
  modeling.
\newblock {\em CoRR}, abs/1412.3555, 2014.

\bibitem{hochreiter1997long}
Sepp Hochreiter and J{\"u}rgen Schmidhuber.
\newblock Long short-term memory.
\newblock {\em Neural computation}, 9(8):1735--1780, 1997.

\bibitem{DBLP:journals/corr/ShazeerMMDLHD17}
Noam Shazeer, Azalia Mirhoseini, Krzysztof Maziarz, Andy Davis, Quoc~V. Le,
  Geoffrey~E. Hinton, and Jeff Dean.
\newblock Outrageously large neural networks: The sparsely-gated
  mixture-of-experts layer.
\newblock {\em CoRR}, abs/1701.06538, 2017.

\bibitem{DBLP:journals/corr/KimDHR17}
Yoon Kim, Carl Denton, Luong Hoang, and Alexander~M. Rush.
\newblock Structured attention networks.
\newblock {\em CoRR}, abs/1702.00887, 2017.

\bibitem{parikh2016decomposable}
Ankur~P Parikh, Oscar T{\"a}ckstr{\"o}m, Dipanjan Das, and Jakob Uszkoreit.
\newblock A decomposable attention model for natural language inference.
\newblock {\em arXiv preprint arXiv:1606.01933}, 2016.

\bibitem{DBLP:journals/corr/LuongPM15}
Minh{-}Thang Luong, Hieu Pham, and Christopher~D. Manning.
\newblock Effective approaches to attention-based neural machine translation.
\newblock {\em CoRR}, abs/1508.04025, 2015.

\bibitem{DBLP:journals/corr/ChorowskiBSCB15}
Jan Chorowski, Dzmitry Bahdanau, Dmitriy Serdyuk, KyungHyun Cho, and Yoshua
  Bengio.
\newblock Attention-based models for speech recognition.
\newblock {\em CoRR}, abs/1506.07503, 2015.

\bibitem{DBLP:journals/corr/SukhbaatarSWF15}
Sainbayar Sukhbaatar, Arthur Szlam, Jason Weston, and Rob Fergus.
\newblock Weakly supervised memory networks.
\newblock {\em CoRR}, abs/1503.08895, 2015.

\bibitem{DBLP:journals/corr/GehringAGYD17}
Jonas Gehring, Michael Auli, David Grangier, Denis Yarats, and Yann~N. Dauphin.
\newblock Convolutional sequence to sequence learning.
\newblock {\em CoRR}, abs/1705.03122, 2017.

\bibitem{hochreiter2001gradient}
Sepp Hochreiter, Yoshua Bengio, Paolo Frasconi, J{\"u}rgen Schmidhuber, et~al.
\newblock Gradient flow in recurrent nets: the difficulty of learning long-term
  dependencies.

\bibitem{NIPS2016_6295}
\L~ukasz Kaiser and Samy Bengio.
\newblock Can active memory replace attention?
\newblock In D.~D. Lee, M.~Sugiyama, U.~V. Luxburg, I.~Guyon, and R.~Garnett,
  editors, {\em Advances in Neural Information Processing Systems 29}, pages
  3781--3789. Curran Associates, Inc., 2016.

\bibitem{DBLP:journals/corr/KalchbrennerESO16}
Nal Kalchbrenner, Lasse Espeholt, Karen Simonyan, A{\"{a}}ron van~den Oord,
  Alex Graves, and Koray Kavukcuoglu.
\newblock Neural machine translation in linear time.
\newblock {\em CoRR}, abs/1610.10099, 2016.

\bibitem{nips}
Ashish Vaswani, Noam Shazeer, Niki Parmar, Jakob Uszkoreit, Llion Jones,
  Aidan~N Gomez, \L~ukasz Kaiser, and Illia Polosukhin.
\newblock Attention is all you need.
\newblock In {\em Advances in Neural Information Processing Systems 30}, pages
  5998--6008. Curran Associates, Inc., 2017.

\bibitem{DBLP:journals/corr/KaiserGC17}
Lukasz Kaiser, Aidan~N. Gomez, and Fran{\c{c}}ois Chollet.
\newblock Depthwise separable convolutions for neural machine translation.
\newblock {\em CoRR}, abs/1706.03059, 2017.

\bibitem{DBLP:journals/corr/Chollet16a}
Fran{\c{c}}ois Chollet.
\newblock Xception: Deep learning with depthwise separable convolutions.
\newblock {\em CoRR}, abs/1610.02357, 2016.

\bibitem{DBLP:journals/corr/abs-1708-00993}
Jan Niehues and Eunah Cho.
\newblock Exploiting linguistic resources for neural machine translation using
  multi-task learning.
\newblock {\em CoRR}, abs/1708.00993, 2017.

\bibitem{li-etal-2018-named}
Zhongwei Li, Xuancong Wang, Ai~Ti Aw, Eng~Siong Chng, and Haizhou Li.
\newblock Named-entity tagging and domain adaptation for better customized
  translation.
\newblock In {\em Proceedings of the Seventh Named Entities Workshop}, pages
  41--46, Melbourne, Australia, July 2018. Association for Computational
  Linguistics.

\bibitem{DBLP:journals/corr/abs-1901-10430}
Felix Wu, Angela Fan, Alexei Baevski, Yann~N. Dauphin, and Michael Auli.
\newblock Pay less attention with lightweight and dynamic convolutions.
\newblock {\em CoRR}, abs/1901.10430, 2019.

\bibitem{he2016deep}
Kaiming He, Xiangyu Zhang, Shaoqing Ren, and Jian Sun.
\newblock Deep residual learning for image recognition.
\newblock In {\em Proceedings of the IEEE conference on computer vision and
  pattern recognition}, pages 770--778, 2016.

\bibitem{ba2016layer}
Jimmy~Lei Ba, Jamie~Ryan Kiros, and Geoffrey~E. Hinton.
\newblock Layer normalization, 2016.

\bibitem{DBLP:journals/corr/YuK15}
Fisher Yu and Vladlen Koltun.
\newblock Multi-scale context aggregation by dilated convolutions.
\newblock In Yoshua Bengio and Yann LeCun, editors, {\em 4th International
  Conference on Learning Representations, {ICLR} 2016, San Juan, Puerto Rico,
  May 2-4, 2016, Conference Track Proceedings}, 2016.

\bibitem{DBLP:journals/corr/KingmaB14}
Diederik~P. Kingma and Jimmy Ba.
\newblock Adam: {A} method for stochastic optimization.
\newblock In Yoshua Bengio and Yann LeCun, editors, {\em 3rd International
  Conference on Learning Representations, {ICLR} 2015, San Diego, CA, USA, May
  7-9, 2015, Conference Track Proceedings}, 2015.

\bibitem{DBLP:journals/corr/BritzGLL17}
Denny Britz, Anna Goldie, Minh{-}Thang Luong, and Quoc~V. Le.
\newblock Massive exploration of neural machine translation architectures.
\newblock {\em CoRR}, abs/1703.03906, 2017.

\bibitem{JMLR:v15:srivastava14a}
Nitish Srivastava, Geoffrey Hinton, Alex Krizhevsky, Ilya Sutskever, and Ruslan
  Salakhutdinov.
\newblock Dropout: A simple way to prevent neural networks from overfitting.
\newblock {\em Journal of Machine Learning Research}, 15:1929--1958, 2014.

\bibitem{wan2013regularization}
Li~Wan, Matthew Zeiler, Sixin Zhang, Yann Le~Cun, and Rob Fergus.
\newblock Regularization of neural networks using dropconnect.
\newblock In {\em International conference on machine learning}, pages
  1058--1066, 2013.

\bibitem{ott2018scaling}
Myle Ott, Sergey Edunov, David Grangier, and Michael Auli.
\newblock Scaling neural machine translation.
\newblock {\em arXiv preprint arXiv:1806.00187}, 2018.

\bibitem{wu2016google}
Yonghui Wu, Mike Schuster, Zhifeng Chen, Quoc~V Le, Mohammad Norouzi, Wolfgang
  Macherey, Maxim Krikun, Yuan Cao, Qin Gao, Klaus Macherey, et~al.
\newblock Google's neural machine translation system: Bridging the gap between
  human and machine translation.
\newblock {\em arXiv preprint arXiv:1609.08144}, 2016.

\bibitem{nguyen2019data}
Xuan-Phi Nguyen, Shafiq Joty, Wu~Kui, and Ai~Ti Aw.
\newblock Data diversification: An elegant strategy for neural machine
  translation, 2019.

\bibitem{zhao2019muse}
Guangxiang Zhao, Xu~Sun, Jingjing Xu, Zhiyuan Zhang, and Liangchen Luo.
\newblock Muse: Parallel multi-scale attention for sequence to sequence
  learning, 2019.

\bibitem{DBLP:journals/corr/ZhouCWLX16}
Jie Zhou, Ying Cao, Xuguang Wang, Peng Li, and Wei Xu.
\newblock Deep recurrent models with fast-forward connections for neural
  machine translation.
\newblock {\em CoRR}, abs/1606.04199, 2016.

\bibitem{DBLP:journals/corr/SzegedyVISW15}
Christian Szegedy, Vincent Vanhoucke, Sergey Ioffe, Jonathon Shlens, and
  Zbigniew Wojna.
\newblock Rethinking the inception architecture for computer vision.
\newblock {\em CoRR}, abs/1512.00567, 2015.

\bibitem{vinyals2015grammar}
Oriol Vinyals, {\L}ukasz Kaiser, Terry Koo, Slav Petrov, Ilya Sutskever, and
  Geoffrey Hinton.
\newblock Grammar as a foreign language.
\newblock In {\em Advances in neural information processing systems}, pages
  2773--2781, 2015.

\bibitem{press2016using}
Ofir Press and Lior Wolf.
\newblock Using the output embedding to improve language models.
\newblock {\em arXiv preprint arXiv:1608.05859}, 2016.

\bibitem{petrov-etal-2006-learning}
Slav Petrov, Leon Barrett, Romain Thibaux, and Dan Klein.
\newblock Learning accurate, compact, and interpretable tree annotation.
\newblock In {\em Proceedings of the 21st International Conference on
  Computational Linguistics and 44th Annual Meeting of the Association for
  Computational Linguistics}, pages 433--440, Sydney, Australia, July 2006.
  Association for Computational Linguistics.

\end{thebibliography}
\nocite{DBLP:journals/corr/ZhouCWLX16}
\nocite{DBLP:journals/corr/SzegedyVISW15}
\nocite{vinyals2015grammar}
\nocite{press2016using}
\nocite{petrov-etal-2006-learning}


\clearpage
\end{document}